\documentclass[letterpaper, 10 pt, conference]{ieeeconf}  

\IEEEoverridecommandlockouts                              

\usepackage{color}
\usepackage{graphicx}
\usepackage{float}
\usepackage{amsmath}
\usepackage{adjustbox}

\usepackage{soul}
\usepackage{verbatim}

\title{\LARGE \bf
Assistive Soft Robotic Glove with Ruffles Enhanced Textile Actuators
}

\author{Cem Suulker$^{1}$, and Kaspar Althoefer$^{1}$%
\thanks{$^{1}$Authors are with the Centre for Advanced Robotics, at the School of Engineering and Materials Science, Queen Mary University of London, United Kingdom.
        {\tt\footnotesize c.suulker@qmul.ac.uk}}%
}

\begin{document}

\maketitle
\thispagestyle{empty}
\pagestyle{empty}

\begin{abstract}

Hand-wearable robots, specifically exoskeletons, are designed to aid hands in daily activities, playing a crucial role in post-stroke rehabilitation and assisting the elderly. Our contribution to this field is a textile robotic glove with integrated actuators. These actuators, powered by pneumatic pressure, guide the user's hand to a desired position. Crafted from textile materials, our soft robotic glove prioritizes safety, lightweight construction, and user comfort. Utilizing the ruffles technique, integrated actuators guarantee high performance in blocking force and bending effectiveness. Additionally, we present a participant study confirming the effectiveness of our robotic device.

\end{abstract}

\section{Introduction}

The field of soft robotics is growing rapidly, and innovative elastic materials are replacing heavy metallic links while soft inflatable actuators are taking the place of electromechanical rotational motors. The use of flexible textile materials in human-robot interaction has also been shown to offer attractive design options due to their safe nature \cite{suulker2022comparison}.

The creation of soft robots and actuators often involves the use of various materials and methods from the clothing industry. One crucial variable is the stretch quality of the fabric material. Knitted fabrics are commonly used for their stretchiness, but they often stretch in all directions, which is usually not desirable. Woven fabrics, are normally known as non-stretchy. But integrating elastane yarn into the weft, makes them more stretch in one direction than the other. This quality makes them more suitable for creating actuators. They are also more durable than knitted ones. While using coating can make these structures airtight, it also eliminates the material's ability to stretch.

Another approach to creating stretch in soft robotic structures involves increasing the material density using methods such as pleats and ruffles. The use of pleats in soft robotics has been extensively researched and is considered an effective method \cite{cappello2018assisting}. However, the use of ruffles has not been researched in the community of soft robotics. Ruffles are often used in the neck area of clothing to gather the fabric material in a dense way. From a soft robotics perspective, this gathered dense material can be inflated to create high elongation.

Various types of elastic bands can be used to create ruffles, but in this extended abstract, we will focus on braided elastic bands. These bands are used for storing energy in soft robotics. However, their full potential is realized when they are integrated into the fabric structure using the ruffles method. In this extended abstract, we will showcase a soft robotic glove that utilizes elastic bands to enhance the system's performance.

\begin{figure}[t]
  \centering
  \includegraphics[width=1\linewidth]{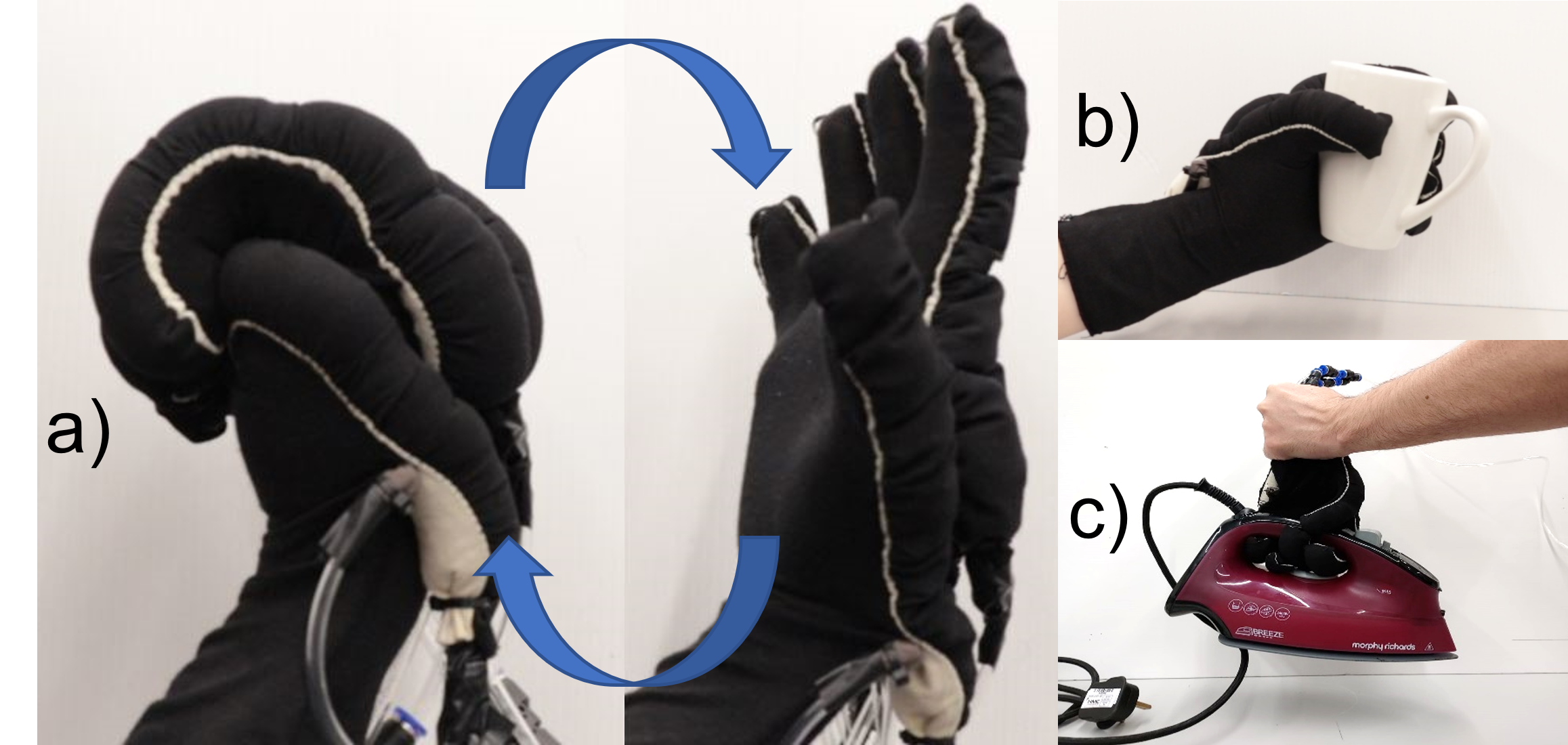}
  \caption{The soft robotic glove prototype in the action of a) closing and opening of the hand, b) grasp assistance, c) lifting an iron (1.3 kg) without a hand in it.}
  \label{fig1}
\end{figure}

\section{Ruffles Enhanced Bending Actuators} 

To create bending of an inflatable textile actuator, an imbalance between two layers of fabric must be achieved. This imbalance is created by using pleating techniques \cite{cappello2018assisting}, excess material of one layer is folded and stitched onto the other, and unfolds when inflated. This approach, however, has potential drawbacks, e.g. when the space between the layers is too small for the pleats to unfold (which, for finger sized designs, can become imminent). Another technique to create such imbalance is the use of different types of fabric with varying elasticity. This way, one layer stretches more when actuated and the structure will bend towards the less elastic one. Integrating braided elastic band to the stretch fabric option with the ruffles technique is proven more efficient in terms of blocking force and bending angle capabilities \cite{suulker2022soft}.

\subsubsection{Materials} 

To create this textile actuator the two different textile properties layers are selected. Bottom: a plain cotton weave for the (light fabric in Fig. 1). Top: a cotton mix with elastane yarn integrated to the weft of the fabric (dark fabric in Fig. \ref{fig1}).

To enhance a fabric's stretch behavior and create the desired material imbalance between layers, an additional support material is used that is integrated when assembling the actuators: a braided elastic band, also called elastics, commonly used in clothing to create ruffles or elastic waistbands. It consists of braided polyester and a small part of thin rubber, making it durable and extremely stretchy.

\subsubsection{Fabrication}

First, the mono-directional stretch black fabric was cut 80\% longer than the cotton bottom layer fabric. The elastic band was integrated on the side seam between the top and bottom layer, first stitched onto the top layer while being stretched, and then sewn onto the bottom layer in a relaxed state. This enabled the top layer to reach an excess length of 180\% of the bottom layer. The actuator is equipped with a latex bladder to ensure air tightness.

\begin{figure}[h]
\centering
  \includegraphics[width=0.9\linewidth]{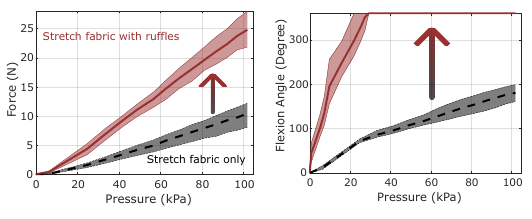}
  \caption{Blocking force output and flexion angle versus pressure graphs for stretch fabric actuator, and elastic band integrated actuator. Integration of the elastic band significantly boosts the performance.}
  \label{fig2}
\end{figure}

\subsubsection{Results}
Two important parameters for soft bending actuators for wearables are flexion angle and blocking force \cite{suulker2022fabric}. In example, for rehabilitation or assistive hand exoskeletons it is imperative that each finger is unrestricted in relation to its maximum angle, and actuators should apply 10-15 N blocking force to the fingers\cite{takahashi2008robot}. 

In Fig. 2 the elastic band integration boosts both of these critical parameters for the actuator. The force capability increases approximately from 10 N to 25 N, and the maximum bending angle increases from approximately 180 degrees to 360 degrees.

\section{Assessment of the Assistive Glove}


The existing literature offers established methodologies for evaluating the effectiveness of hand-wearable assistive devices through user studies, utilizing tests such as the Jebsen Taylor Hand Function Test \cite{jebsen} and the Box and Blocks Test \cite{mathiowetz1985adult}. These assessments are particularly valuable when conducted with individuals whose manual dexterity is compromised in some capacity. However, there are considerable challenges associated with working with such participant groups. In addition to ensuring the safety of the robotic prototype for use by fragile or vulnerable individuals, complex ethical considerations must be addressed.

Given these challenges, many researchers opt to conduct studies with sample groups comprised of relatively "unproblematic" individuals. Nevertheless, this approach introduces its own complexities, particularly concerning trust. Healthy or non-compromised users may inadvertently influence test results due to their natural dexterity. Their ability to grasp and hold objects can either override or assist the actuation of the robotic device. This issue is particularly pertinent with soft robotic devices, which possess high flexibility and relatively low actuation force. Consequently, it is not advisable to work with a sample of healthy or non-compromised individuals.

Instead, to accurately quantify the assistance provided by the robotic device, Electromyography (EMG) sensing should be considered. EMG sensing offers a more objective measure of the user's interaction with the device, allowing researchers to evaluate the device's effectiveness in providing assistance independently of the user's inherent dexterity.

\begin{figure}[h]
\centering
  \includegraphics[width=0.85\linewidth]{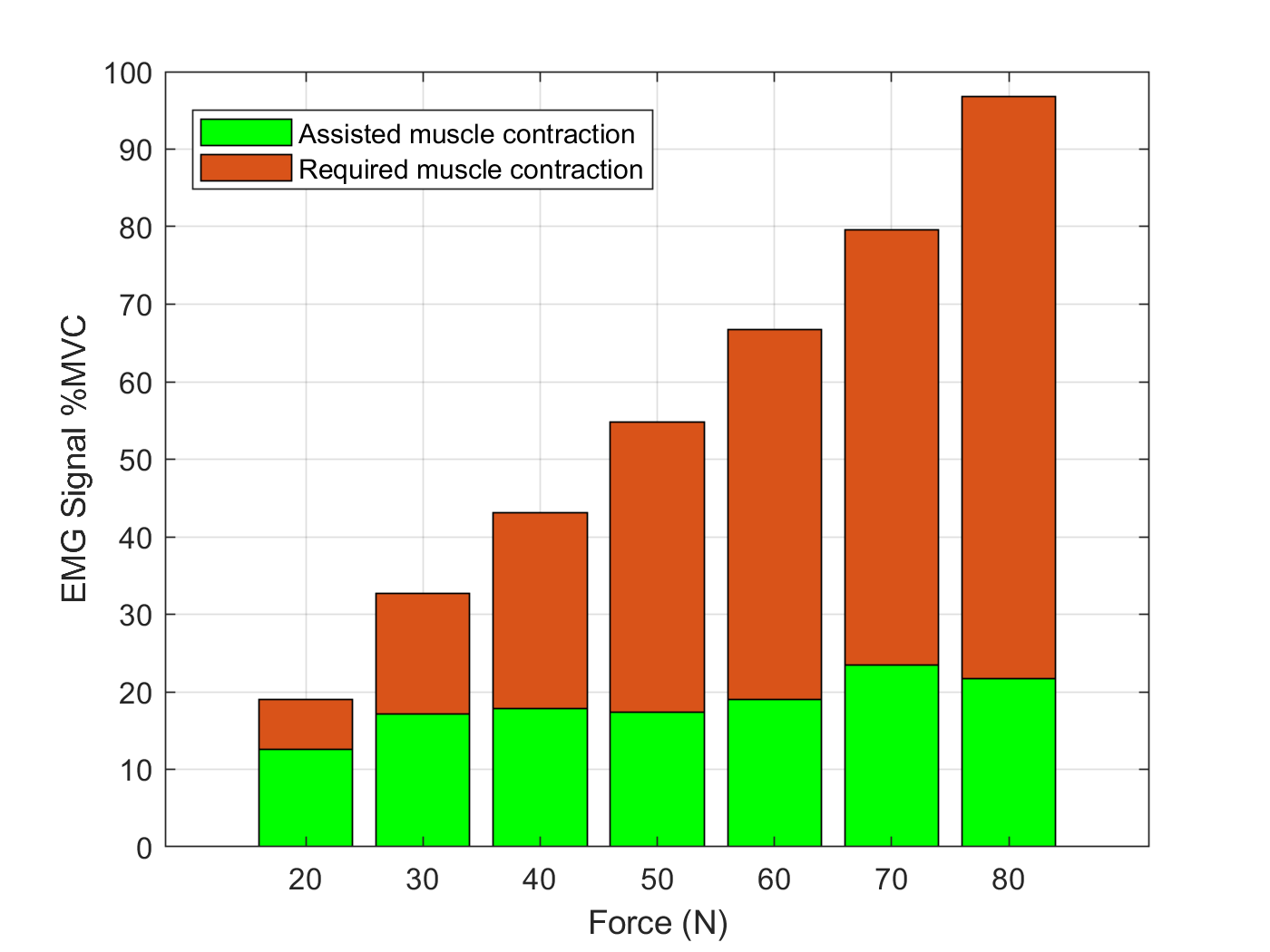}
  \caption{Bar plot that shows muscle contraction assisted by the glove (green), and the muscle contraction needed for the task (red), for different force intensity of tasks.}
  \label{bar}
\end{figure}

In our latest study, we propose a novel method to evaluate wearable devices with healthy participants using EMG sensing. We then apply this method to assess the effectiveness of our soft robotic assistive prototype glove. Our results demonstrate that our exoskeleton significantly improves participants' grasping ability. We record EMG signals as participants perform a task involving varied grasp force applied to an object with and without the assistance of the robotic system. The benefits of the assistive device are quantified using linear mixed-effects models. As our work is currently under review, we provide only a preview of the results in Figure \ref{bar}. For an extended version of this two page abstract please visit \cite{10608402}.



\bibliographystyle{IEEEtran}
\bibliography{references.bib}

\end{document}